\documentclass[a4paper,fleqn]{cas-dc}

\usepackage[numbers]{natbib}

\def\tsc#1{\csdef{#1}{\textsc{\lowercase{#1}}\xspace}}
\tsc{WGM}
\tsc{QE}
\newcommand{\etal}{\textit{ et al.}}
\newcommand{\etc}{\textit{ etc}}

\begin{document}
\let\WriteBookmarks\relax
\def\floatpagepagefraction{1}
\def\textpagefraction{.001}

\shorttitle{A new baseline for edge detection: Make Encoder-Decoder great again}    

\shortauthors{Yachuan Li et al.}  

\title [mode = title]{A New Baseline for Edge Detection: make encoder-decoder great again}  



%

\author[1]{Yachuan Li}[orcid=0000-0002-4516-5576]
\cormark[1]
\ead{liyachuan@s.upc.edu.cn}

\credit{Conceptualization, Writing—original draft}

\author[2]{Xavier Soria Poma}[orcid=0000-0003-2997-2439]
\ead{xavier.soria@espoch.edu.ec}
\credit{Data curation, Writing—review and editing}

\author[3]{Yongke  Xi}
\ead{xiyk_lrving@163.com}
\credit{Data curation}

\author[1]{GUANLIN LI}
\ead{liguanlin@s.upc.edu.cn}
\credit{Formal analysis}

\author[1]{Chaozhi Yang}
\ead{yang.chaozhi@foxmail.com}
\credit{Investigation, Validation}

\author[1]{Qian Xiao}
\ead{xiaoqian@s.upc.edu.cn}
\credit{Investigation, Validation}

\author[1]{Yun Bai}
\ead{baiyun@s.upc.edu.cn}
\credit{Investigation, Validation}

\author[1]{Zongmin LI}[orcid=0000-0003-4785-791X]
\credit{Conceptualization, Resources, Supervision}
\ead{lizongmin@upc.edu.cn}
\cormark[1]

\address[1]{China University of Petroleum (East China), Qingdao 266500, China}
\address[2]{ Polytechnic School of Chimborazo (ESPOCH), EspochAI, Morona-Santiago, 14060, Ecuador}
\address[3]{SHANDONG HI-SPEED INFORMATION GROUP CO,LTD, Jinan 250014,China }

\cortext[1]{Corresponding author}

\makeatletter\def\Hy@Warning#1{}\makeatother
\fntext[1]{
Thanks to National key r\&d program (Grant no. 2019YFF0301800), 
National Natural Science Foundation of China (Grant no. 61379106), 
and the Shandong Provincial Natural Science Foundation (Grant nos.ZR2013FM036, ZR2015FM011) 
for funding.}


\begin{abstract}
The performance of deep learning based edge detector has far exceeded that of humans, but the huge computational cost and complex training strategy hinder its further development and application. In this paper, we eliminate these complexities with a vanilla encoder-decoder based detector. Firstly, we design a bilateral encoder to decouple the extraction process of location features and semantic features. Since the location branch no longer provides cues for the semantic branch, the richness of features can be further compressed, which is the key to make our model more compact.
We propose a cascaded feature fusion decoder, where the location features are progressively refined by semantic features. The refined location features are the only basis for generating the edge map. The coarse original location features and semantic features are avoided from direct contact with the final result. So the noise in the location features and the location error in the semantic features can be suppressed in the generated edge map.
The proposed New Baseline for Edge Detection (NBED) achieves superior performance consistently across multiple edge detection benchmarks, even compared with those methods with huge computational cost and complex training strategy. 
The ODS of NBED on BSDS500 is 0.838, achieving state-of-the-art performance.
Our study shows that what really matters in the current edge detection is high-quality features, and we can make the encoder-decoder based detector great again even without complex training strategies and huge computational cost. 
The code is available at https://github.com/Li-yachuan/NBED.

\textbf{Note:} This work has been accepted for publication in \textit{Signal Processing: Image Communication}. A definitive version is available at: https://doi.org/10.1016/j.image.2026.117485

\end{abstract}


\begin{highlights}
\item Huge computational cost and complex strategy hinder the progress of edge detection
\item A vanilla and pure encoder-decoder based edge detection baseline is needed
\item NBED decouples the extraction process of location features and semantic features
\item Cascaded high-resolution feature decoder can suppress the noise and the location error
\item Powerful encoder-decoders can disable the feature augmentation and fusion modules
\end{highlights}


\maketitle


\section{Introduction}

Edge detection captures object boundaries and salient edges in images, which can keep the important semantic information in the image and discard the irrelevant noise such as color and texture.
As a result, edge detection is one of the most important basic tasks of computer vision and is widely used in many downstream tasks, such as image segmentation~\cite{liu2021dance}, depth map prediction~\cite{zhu2020edge} and salient detection~\cite{zhao2019egnet}.

Early researches on edge detection~\cite{davis1975survey,sharifi2002classified,sobel19683x3,canny1986computational,prewitt1970object} mainly rely on local features, but the local features of edges and texture noise are highly similar and hard to distinguish. Therefore, the performance of early edge detectors is hardly satisfactory. 
With the application of deep learning in edge detection~\cite{xie2017holistically,liu2019richer,he2022bdcn,huan2021unmixing,soria2022ldc}, the performance of detectors has been improved unprecedented, far surpassing the local feature based methods and even surpassing human beings.
Deep learning based edge detectors have become the mainstream methods for current edge detection.

As the pioneer of contemporary edge detection, HED first introduces the deep supervision mechanism to edge detection and learns multi-scale predictions holistically, which dominates the development of edge detection at that time~\cite{liu2019richer,he2022bdcn,huan2021unmixing}.
To keep the lightweight of the model, these methods use extremely simple decoders to fuse and compress the multi-scale information into single-channel edges. 
With the development of deep learning, the encoder and decoder are gradually upgraded iteratively.
Pu\etal~\cite{pu2022edter} replace the encoder with Transformer.
Deng\etal~\cite{deng2018learning} introduce encoder-decoder with symmetric structure into edge detection.
But this encoder-decoder structure has remained the same. The encoder extracts multi-scale features including location information and semantic information, and the decoder fuses these features to obtain the final edge map.

In recent years, the development of this structure seems to reach a bottleneck and researchers turn to focus on the optimization of data processing and training strategies~\cite{zhou2023uaed,ye2024diffusionedge,li2023kded}.
These methods further improve the performance of edge detectors by using complex data processing and training strategies such as data distillation~\cite{li2023kded}, multi-stage training~\cite{pu2022edter}, multi-step inference~\cite{ye2024diffusionedge}, auxiliary model~\cite{fu2023practical},\etc.

However, there are some new problems worth exploring behind the booming development of edge detection. 
\textbf{Firstly}, there is a conflict between the effectiveness and efficiency of the model. To improve the performance of the model, the size and computational cost of the model is expanding.
\textbf{Secondly}, The location error in the semantic features and the texture noise in the location features can negatively affect the generated edge map.
\textbf{Thirdly}, 
the complex training processing and training strategies make edge detection an extremely complex task, which hinders the application and promotion of edge detection methods.

Having in mind those issues, we try to abandon the complex strategies and design an effective and compact edge detector with vanilla encoder-decoder. This bilateral encoder with decoupled location features and semantic features. Location features no longer need to provide rich information cues for semantic features, so the channels of location branch can be further compressed. 
The semantic branch focuses on extracting semantic information rather than location information, so features keep rich channels and reduce resolution. The semantic branch is a CNN-Transformer hybrid model where Transformer is only applied on the features with lowest resolution. 
This can make the semantic branch keep the balance between performance and efficiency.
Role of the decoder to fuse the multi-scale information and obtain the final edge map. We design a cascaded feature fusion network, in which the high-resolution location features are gradually fused with the semantic features.
The edge map is generated only by the refined location features. 
Neither the coarse location features with texture noise nor the low-resolution semantic features with localization error directly connect to the final edge map. It can effectively improve the quality of edge map.

Without any other additional data processing and the training strategy, only rely on the simple structure, our method achieved competitive results on benchmarks, which achieve state-of-the-art performance on BSDS500 and BIPED. 
In addition, we verify the effectiveness of some existing feature enhancement and feature fusion modules, and the experimental results show that the effectiveness of these modules is greatly discounted when the encoder is powerful enough.

\section{Related Work}
The structures of early edge detectors are relatively simple, and the edge is judged by the change of local texture and color information. The accuracy has been surpassed by the methods based on deep learning. In addition to being used in a very few scenarios such as the lack of supervision information, few people pay attention to them now, so we mainly discuss the edge detection based on deep learning. Unless otherwise specified, edge detection mentioned below generally refers to deep learning based edge detection.

\subsection{The Encoder of Edge Detection}

The pioneer of contemporary edge detection is HED~\cite{xie2017holistically}, which extracts features using VGG~\cite{simonyan2014very}, the state-of-the-art feature extraction network at the time, and fuses multi-scale features with several convolutions and upsampling to get the final edge. Although rough, HED fuses local and global information for the first time, which makes it perform much better than other detectors based on local information and achieves the highest accuracy. Inspired by HED, a series of excellent edge detectors~\cite{liu2019richer,he2022bdcn,huan2021unmixing} have emerged. For fair comparison, there is a tacit agreement to use VGG to extract features and make the decoder as lightweight as possible. 

With the development of deep learning, stubborn selection of VGG to extract features is not only not conducive to further improvement of model accuracy, but also difficult to flexibly adjust the size of the model. Therefore, researchers begin to explore new encoders.
On the one hand, they explore more powerful encoders to obtain higher accuracy edges. Both RCF~\cite{liu2019richer} and BDCN~\cite{he2022bdcn} try to directly use ResNet~\cite{he2016deep}, a more advanced encoder to improve the performance of their models. EDTER~\cite{pu2022edter} introduces Transformer into edge detection for the first time, which brings huge computation, but also improves the accuracy by leaps and bounds. The important innovation of DiffusionEdge~\cite{ye2024diffusionedge} is to introduce Diffusion to edge detection, but its encoder is also updated with the latest SWin-Transformer.
On the other hand, more efficient encoders are designed to obtain lightweight models, FINED~\cite{wibisono2021fined} only uses the first few convolutions of VGG to avoid a lot of computation, and LDC~\cite{soria2022ldc} utilizes channel separable convolutions and a smaller number of channels to design a more lightweight encoder. PiDiNet~\cite{su2021pixel} combines channel separable convolutions with handcrafted features to obtain high-quality edges with a lighter weight model.

Edge detection requires location information and semantic information. In previous works~\cite{liu2019richer,pu2022edter,zhou2023uaed}, both are extracted by the same encoder. This may seem to simplify the design of the model, but shallow features need to maintain high resolution to ensure the accuracy of location information, while keep a sufficient number of channels to provide rich information cues for deep features. This leads to an inevitable increase in model parameters and computational cost.
This problem is exacerbated after Transformer is introduced to edge detection~\cite{pu2022edter,ye2024diffusionedge}, which results in tens of times more parameters compared to previous CNN-based methods~\cite{liu2019richer,he2022bdcn}.

\subsection{The Decoder of Edge Detection}
To the best of our knowledge, the two most commonly used decoders in edge detection are the HED-style decoder~\cite{xie2017holistically,he2022bdcn,xuan2022fcl} and the UNet-style decoder~\cite{deng2018learning,deng2020deep}.
The HED-style decoder fuses and compresses the multi-scale features into a single-channel edge map using only simple upsampling and $1\times1$ convolution.
The HED-style decoder can obtain the location information in the shallow features and the semantic information in the deep features at the same time.  
However, shallow features will contain a lot of noise due to the lack of semantic information, and semantic features will introduce location errors due to excessive downsampling. The existence of these two kinds of errors seriously affects the quality of the generated edge map.
The UNet-style decoder gradually fuses high-level semantic information with shallow features, and compresses the fused shallowest features into a single-channel edge map.
Therefore, the adverse effects of location errors in the semantic information on the edge map can be alleviated. However, the close contact between shallow features and the final result inevitably mixes noise into the generated edge map.


\subsection{Training Strategies for edge detection}
\label{sec-training-strategy}
With the development of edge detection models gradually becoming more mature, researchers have gradually shifted their attention to data processing and training strategies.

UAED~\cite{zhou2023uaed} studies the subjectivity and ambiguity of different annotations by estimating the uncertainty of the data. This not only requires the dataset to have multiple sets of labels, but also requires additional processing of these labels.
KDED~\cite{li2023kded} and PEdge~\cite{fu2023practical} eliminate the adverse effects caused by subjectivity and ambiguity of annotations by distillation and self-distillation, which make the training process more complex.
PiDiNet~\cite{su2021pixel} and PiDiNeXt~\cite{li2023pidinext} achieve efficient inference through reparameterization, and EDTER~\cite{pu2022edter} achieves global context modeling and local refinement through multi-stage training, respectively. DiffusionEdge~\cite{ye2024diffusionedge} introduces Diffusion to the edge detection task and transforms edge detection into a multi-step denoising process.

These methods effectively improve the performance of edge detection, but also increase the difficulty of training and deployment of the models, which hinder the development and application of edge detection methods. In theory, if the model is powerful enough, random errors in the labels can be learned by the model. 
Therefore, we endeavour to design an encoder-decoder that is powerful enough to perform the edge detection task simply and elegantly without these complex strategies.

\section{A NEW BASELINE FOR EDGE DETECTION}

\begin{figure*}[ht]
  \centering
   \includegraphics[width=\linewidth]{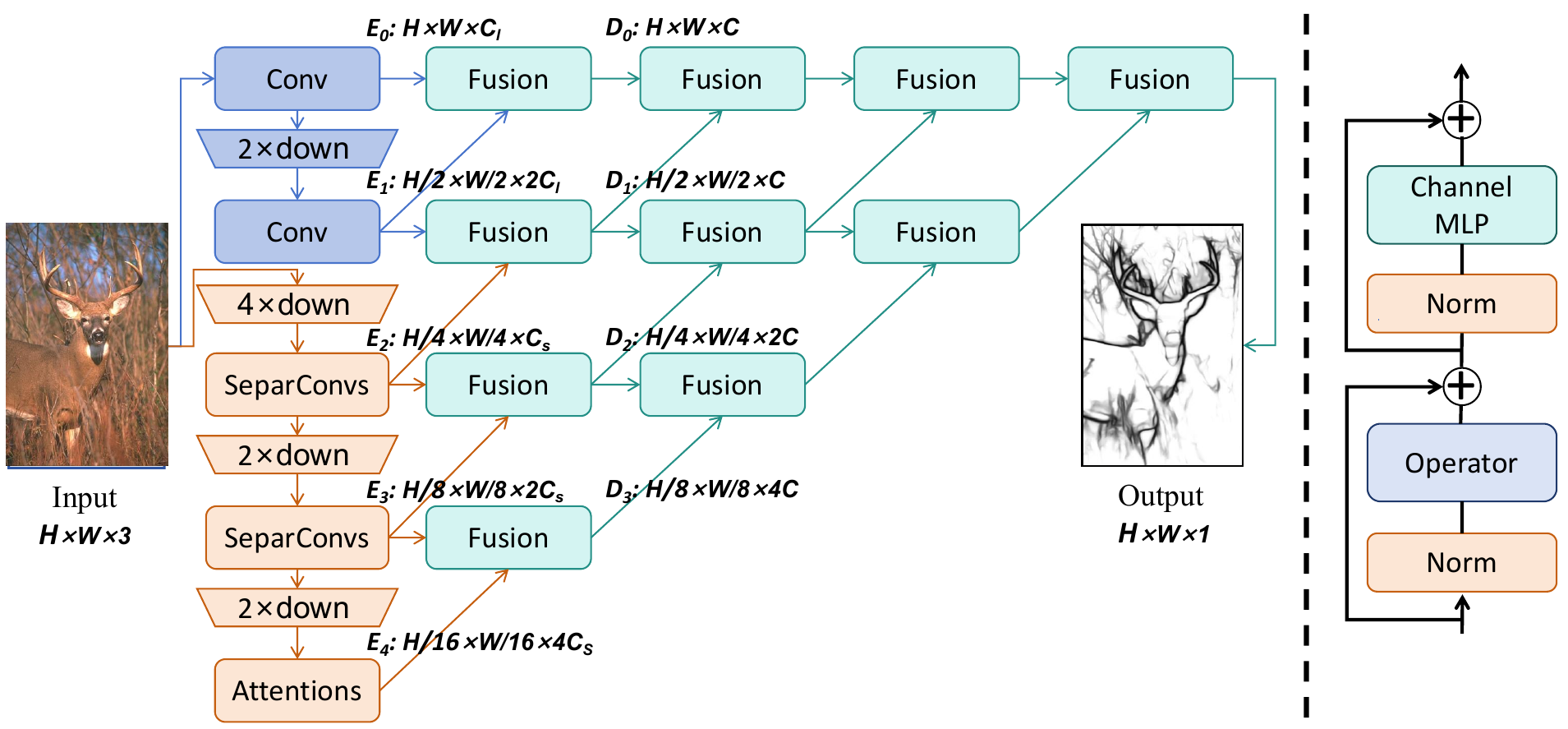}
   \hspace*{6cm}(a) Overall framework
    \hspace*{6cm}(b) Meta block 

   \caption{(a) Overall framework of NBED. \textit{\textbf{SeparConvs}} and \textbf{\textit{Attentions}} contain multiple \textit{\textbf{SeparConv}} and \textit{\textbf{Attention}} modules, respectively.
   \textit{\textbf{n$\times$ down}} means n $\times$ downsampling. $E_i$: Feature size at stage \textit{i} of the encoder, $D_i$: Feature size at stage \textit{i} of the decoder.
   (b) \textbf{\textit{Meta block}}. \textbf{\textit{SeparConv}} and \textbf{\textit{Attention}} in (a) are composed of \textbf{\textit{Meta block}} and different \textbf{\textit{Operator}}s.
   }
   \label{fig:arch}
\end{figure*}

Our original intention is to avoid various complex training strategies, as mentioned in  Sec.~\ref{sec-training-strategy}, and train a vanilla encoder-decoder, so we put our effort into the model architecture. The Overall framework of NBED is shown in Fig.~\ref{fig:arch}, which is an encoder-decoder structure as a whole. 
Given an image $X \in \mathbb{R}^{H \times W \times 3}$,
we first extract location feature and semantic feature by inputting $X$ into the location feature encoder and the semantic feature encoder respectively. The two encoders are shown in light blue and light orange respectively. And the extracted multi-scale features are uniformly fed to the decoder, the light green module in the figure. After aggregating and compressing the output of the decoder, the final edge map $\hat Y \in \mathbb{R}^{H \times W \times 1}$ can be obtained.

We will detail the location feature encoder in Sec.~\ref{sec:lfe}, the semantic feature encoder in Sec.~\ref{sec:sfe}, the cascaded feature fusion decoder in Sec.~\ref{sec:cffd} and network training in Sec.~\ref{sec:nt}.

\subsection{Location Feature Encoder}
\label{sec:lfe}

In previous works~\cite{liu2019richer,he2022bdcn,xuan2022fcl}, multi-scale features are usually extracted by the same encoder. As a result, the shallow module is not only used to generate location features, but also used to provide cues for higher-level modules.
For the shallow feature $L_i\in \mathbb{R}^{H_i \times W_i \times C_i}$, to ensure the accuracy of the location information, the resolution $H_i \times W_i$ must be large enough. At the same time, the number of channels $C_i$ cannot be reduced to keep rich cues for semantic information. 
So $L_i$ is usually a feature with heavy parameters.
In addition, the problem of gradient instability forces researchers to employ additional processing such as regularization~\cite{ioffe2015batch} and residual connection~\cite{he2016deep}, even in shallow modules. It makes the computation of shallow features more complex.

We propose an independent location feature encoder, a shallow network.
The location feature encoder contains two vanilla Convolution blocks, and each vanilla convolution block consists of a $3\times 3$ convolution module and a ReLU module. 
To ensure the accuracy of the location information, the feature resolution in the first vanilla Convolution block is consistent with the input image.
A down-sampling module is inserted between the two convolution blocks to generate two-scale location features, which can reduce the scale gap between location features and semantic features. Consistent with VGG16~\cite{simonyan2014very}, the number of channels doubles when the feature resolution is halved.
Since providing cues for semantic information is no longer considered, the features in Location Feature Encoder is compressed to 16 channels, which is much smaller than existing models such as VGG16~\cite{simonyan2014very} (64 channels) and ResNet50~\cite{he2016deep} (64 channels). 
Compared to VGG16 and ResNet50, our Location Feature Encoder not only has fewer modules, but also has fewer channels per module.
Therefore, even though the features keep high resolution, the computational cost remains low.

In short, our Location Feature Encoder consists of 2 convolution-ReLU blocks with low channel count, which is expected to maintain the accurate location information of edges in the image with less cost.

\subsection{Semantic Feature Encoder}
\label{sec:sfe}
\begin{table}[ht]
\setlength{\tabcolsep}{4pt} 
    \centering
    \begin{tabular}{c|c|c|c|c}
    \toprule
        Stages  & Blocks    & Channels  & Operator  &Downsampling\\
        \hline
        1       & 3         &96         & SeparConv &1/4\\
        2       &12         &192        & SeparConv &1/8\\
        3       &18         &384        & Attention &1/16\\
    \bottomrule
    \end{tabular}
    \caption{Model configurations of Semantic Feature Encoder. \textbf{\textit{SeparConv}}: depthwise separable convolution, \textit{\textbf{Attention}}: Self-Attention. The downsampling rate is relative to the input image.}
    \label{tab:sfe-cfg}
\end{table}

As shown in the light orange module of Fig.~\ref{fig:arch} (a), the semantic feature encoder is divided into three stages, and at the beginning of each stage, the features are downsampled to different scales.
The output of each stage is called semantic features, which together with location features constitute multi-scale features.

Each stage contains multiple meta blocks, and the structure of meta block is shown in Fig.~\ref{fig:arch} (b). 
Meta block consists of two cascaded residual blocks; the first residual block contains a regularization layer and a operator, and the second residual block contains a regularization layer and a channel MLP to finish the information exchange between channels.
The residual connection and regularization layer can ensure that the gradient is smooth during back-propagation. The operator can aggregate local or semantic information. And channel MLP can complete the information interaction between channels.

The cascaded residual block is actually an important component of the Transformer and together with the Self-Attention module contributes to the success of the Transformer~\cite{liu2021swin}.
We intended to use Transformer to capture the semantic features needed for edge detection, but the massive computational cost of Self-Attention (SA) prevents us, even if the image is downsampled before being fed into the semantic feature encoder. Inspired by CAFormer~\cite{yu2023metaformer}, we adopt a CNN-Transformer hybrid model that uses CNN in the first two stages with large resolution and SA only on the low-resolution features in the last stage. 
To further reduce the computational cost, the CNN in the first two stages is replaced by channel separable convolution, which is consistent with CAFormer~\cite{yu2023metaformer}.

In the first stage and second stage, the operator of meta block is channel separable convolution, and in the last stage, the operator of meta block is Self-Attention.
The CNN-Transformer hybrid model is the best solution to balance performance and efficiency.  
The detailed configurations of the semantic feature encoder can be obtained from Table.~\ref{tab:sfe-cfg}.

\subsection{Cascaded Feature Fusion Decoder}
\label{sec:cffd}

The decoder structure is shown in the light blue module in fig.~\ref{fig:arch} (a), where the high-resolution location features are gradually fused with the low-resolution semantic features, and then the channel of the refined location features is compressed to obtain a single channel edge map. 

The process of \textit{j}-th feature fusion for the \textit{i}-th scale feature \textit{F} can be expressed as \(F_i^j = F_i^{j-1} \oplus \phi(F_{i+1}^{j-1})\), where $\phi(\cdot)$ represents adjusting the features by upsampling and channel MLP, and $\oplus$ represents the fusion of two features by concatenation and convolution in the channel dimension. 
Following the principle that low-resolution features have more channels, the feature size of stage \textit{i} is shown by $D_i$ in fig.~\ref{fig:arch}, where $C = 32$.
Cascaded feature fusion decoder has two main advantages:

Firstly,
mitigate the adverse effects of location errors in semantic features.
The encoder obtains richer semantic information by downsampling, but at the same time, it will lose accurate location information. In Cascaded feature fusion decoder, the refined location feature is the only basis for generating the edge map. Avoiding the direct contact between the semantic feature and the edge map can suppress the adverse effects of the location error in the final edge map.

Secondly,
suppress the expression of texture noise in location features.
The location features of the multi-scale feature are only generated by several convolutions, and the receptive fields of these convolutions are relatively small. 
The small receptive field results in the lack of semantic information in the features, and it is difficult to distinguish edges from textures by only relying on local features, so the location features inevitably contain texture noise.
Cascaded feature fusion decoder can gradually refine the location features with semantic features, and the semantic information in the semantic features can suppress the texture noise in the location features.

\subsection{Network Training}
\label{sec:nt}
We adopt the annotator-robust loss proposed in RCF~\cite{liu2019richer} for all generated edge maps. It can be formulated as
\begin{equation}
    \label{eq-wce}
	WCE\left(p_{i}, y_{i}\right)=\left\{
	\begin{array}{cc}
		-\alpha \log \left(p_{i}\right)  &  { if }\ \mathrm{y}_{i}>\eta \\
		-\beta \log \left(1-p_{i}\right) &  { if }\ \mathrm{y}_{i}=0 \\
		0								 &  { otherwise }
	\end{array}\right.
\end{equation}
where \textit{p} denotes the final edge prediction and \textit{y} is the ground truth. $\alpha=\lvert Y_{+} \rvert/\lvert Y \rvert$ and $\beta =\lambda\cdot\lvert Y_{-} \rvert/\lvert Y \rvert$. $\lvert Y_{+} \rvert$ and $\lvert Y_{-} \rvert$ are used to represents the number of edge and non-edge, separately. $\lambda$ controls the weight of positive over negative samples. $\lvert Y \rvert=\lvert Y_{+} \rvert + \lvert Y_{-} \rvert$. Due to the inconsistency of annotations among different annotators, a threshold $\gamma$ is introduced for loss computation \cite{soria2023dexined_ext}.  For pixel \emph{i}, it will be regarded as edge if the ground truth $y_{i}$ is more than $\gamma$, and we define the ground truth $y_{i}=1$. Pixel \textit{i} will be regarded as no-edge if $y_{i}=0$. $\lambda$ controls the weight of edge over non-edge. 

The training process of NBED is quite concise and does not require additional data processing and complex training strategy, which is unique among the current state-of-art edge detectors, so NBED has the potential to become a new baseline.

\section{Experiments}

\subsection{Datasets} 

We evaluate the proposed NBED on three widely used datasets BSDS500, NYUDv2, and BIPED.
\textbf{BSDS500}~\cite{arbelaez2010contour} is by far the most widely used dataset in edge detection, which includes 200 training images, 100 validation images, and 200 test images. Four to nine annotated maps are associated with each image. The final edge ground truth is computed by taking their average. NBED are trained on training and validation images, and tested on the test images. 
\textbf{NYUDv2}~\cite{silberman2012indoor} is an indoor scene semantic segmentation dataset, the edge ground truth is generated from segmentation maps. it contains 1449 groups of carefully annotated RGB and depth images, and each group image has one annotation result. We use 795 images to train the model and evaluate the model on the rest images.
\textbf{BIPED} \cite{poma2020dense, soria2023dexined_ext} contains 250 carefully annotated high-resolution Barcelona Street View images. There are 200 images for training and validation, and 50 images for testing. All images are carefully annotated at single-pixel width by experts in the computer vision field.

The data augmentation strategy during training is shown in Table~\ref{tab:dataAug}. The data augmentation strategy of BSDS is consistent with LPCB~\cite{deng2018learning}. The data augmentation strategy of NYUDv2 follows RCF~\cite{liu2019richer}, with the only difference that the augmentation is randomly cropped to $400\times400$ for batch calculation. The data augmentation strategy of BIPED is consistent with DexiNed~\cite{poma2020dense}.

\begin{table}[ht]
    \centering
    \begin{tabular}{c|c|c}
        \toprule
        Datasets & Augmentation strategies& Resize\\
        \hline
         BSDS& F ($4\times$), R ($25\times$) & $481\times321$\\
         NYUDv2& F ($2\times$), S ($3\times$), R ($4\times$)&$400\times400$\\
         BIPED&F ($2\times$), S ($3\times$), R ($16\times$), G ($3\times$)&$400\times400$\\
         \bottomrule
    \end{tabular}
    \caption{Augmentation strategies adopted on three edge detection benchmarks. \textbf{\textit{F}}: flipping, \textbf{\textit{S}}: scaling, \textbf{\textit{R}}: rotation, \textbf{\textit{C}}: cropping, \textbf{\textit{G}}: gamma correction.}
    \label{tab:dataAug}
\end{table}

\subsection{Implementation details}

We implement our NBED using PyTorch library. All parameters are updated by Adam optimizer. To better obtain the semantic information of images, the semantic feature encoder is pre-trained on ImageNet, and the remaining modules are randomly initialized. The batchsize is 4 and the weight decay is set to 5e-4. For the pretrained module, the learning rate is 1e-5. the rest is set to 1e-4. The threshold $\eta$ is set to 0.3 for BSDS. No $\eta$ is needed for NYUDv2 and BIPED since the images are singly annotated. The hyperparameter $\lambda$ is set to 1.1 for both BSDS500 and BIPED, and 1.3 for NYUDv2.

A standard Non-Maximum Suppression (NMS) is performed to produce the final edge maps before the quantitative evaluation. F-measure is utilized as the quality evaluation standard of the generated edge map:
\begin{equation}
	F_{-} measure=\frac{2\times P\times R}{P+R}
\end{equation}
where \emph{P} represents the accuracy rate and \emph{R} represents the recall rate. 

Due to local correlation, the edges obtained by deep learning-based methods are edge probability maps, even after NMS processing. So we need to choose a threshold to binarize the edges.
There are two options for the threshold used to binarize the edges. One is to find an optimal threshold value for each image (OIS), and another is to use the same optimal threshold for the whole dataset (ODS). 
The maximum tolerance allowed for correct matches between edge predictions and ground truth annotations is set to 0.0075 for BSDS500 and BIPED, and 0.011 for NYUDv2.
For more experimental details, please refer to previous works~\cite{huan2021unmixing,poma2020dense}.

In addition to accuracy, the model size and the inference efficiency of the model are also noteworthy metrics, which can be reflected by the Giga Floating-Point Operations (GFLOPs) of the model. Compared with Frame Per Second (FPS), GFLOPs is not affected by hardware and can objectively reflect the efficiency of the model.

\subsection{Comparison with state of the arts}
\begin{table*}[ht]
    \centering
    \begin{tabular}{c|c|cc|cc|c|c}
    \toprule 
         \multirow{2}*{Methods}& \multirow{2}*{Pub.'Year}& \multicolumn{2}{c|}{SS} & \multicolumn{2}{c|}{MS} &  \multirow{2}*{Param. (M)}&  \multirow{2}*{GFLOPs}\\
         \cline{3-6}
         && ODS & OIS &ODS & OIS & &\\
         \hline
         Canny~\cite{canny1986computational}&  PAMI'86& 0.611     & 0.676  &-  &-&-&- \\
         gPb-UCM~\cite{arbelaez2010contour}&PAMI'10  & 0.729     & 0.755  &-  &-&-&- \\
         SCG~\cite{ren2012discriminatively}& NeurIPS'12 & 0.739     & 0.758  &-  &- &-&-\\
         SE~\cite{dollar2014fast}& PAMI'14 & 0.743     & 0.764 &-  &-&-& -\\
         OEF~\cite{hallman2015oriented} & CVPR'15      & 0.746     & 0.770   &-  &-&-&- \\
         \hline
         DeepEdge~\cite{bertasius2015deepedge}&CVPR'15      & 0.753     & 0.772   & - &-&-&- \\
         DeepContour~\cite{shen2015deepcontour}& CVPR'15      & 0.757     & 0.776  & - & -&-&-\\
         Deep Boundary~\cite{kokkinos2015pushing}&ICLR'15   &0.789&0.811  & 0.803 & 0.820 &-&-\\
         CEDN~\cite{yang2016object}&  CVPR'16      & 0.788     & 0.804  & - & - &- &- \\
         RDS~\cite{liu2016learning}& CVPR'16      & 0.792     & 0.810  & - & - &  - &- \\
         AMH-Net~\cite{xu2017learning}& NeurIPS'17   & 0.798     & 0.829    & - & - &-&-\\
         HED~\cite{xie2017holistically}& PAMI'17      & 0.788     & 0.808  & - &- &14.6&57.5\\
         CED~\cite{wang2017deep} & CVPR'17   &0.803&0.820  &-  & -&21.8&-\\
         LPCB~\cite{deng2018learning}& ECCV'18    &0.800&0.816  &-  &-&-&- \\
         RCF~\cite{liu2019richer}& PAMI'19   &0.798&0.815  & - &-&14.8&75.3 \\
         DSDC~\cite{deng2020deep}& ACMMM'20 &0.802&0.817  & - &-&-&- \\
         LDC~\cite{deng2021learning}& ACMMM'21   &0.799&0.816  & - & -&-&-\\
         PiDiNet~\cite{su2021pixel} & ICCV'21     & 0.789 & 0.803&-&-&0.71&11.6\\
         BDCN~\cite{he2022bdcn}& PAMI'22   &0.806&0.826 & - & -&16.3&103.4\\
         FCL~\cite{xuan2022fcl}& NN'22   &0.807&0.822  & 0.816 & 0.833&16.5&134.4\\
         EDTER~\cite{pu2022edter}&CVPR'22  & 0.824 &0.841  & 0.840 &\textbf{0.858}&468.8&802.3 \\
         UAED~\cite{zhou2023uaed}&  CVPR'23& 0.829  & 0.847 & 0.837 & 0.855&72.5&72.8\\
         DiffusionEdge~\cite{ye2024diffusionedge}& AAAI'24 & 0.834 & 0.848 & - & -&225&494\\
         NBED (ours)& - & \textbf{0.838} & \textbf{0.849} & \textbf{0.845} & \textbf{0.858}&40&68.8\\
    \bottomrule
    \end{tabular}
    \caption{Quantitative results on the BSDS dataset. For fair comparison, we list the single-scale results and multi-scale results generated by models trained with only BSDS data.}
    \label{tab:rsult-bsds}
\end{table*}

\begin{figure*}[ht]
    \centering
    \includegraphics[width=1\linewidth]{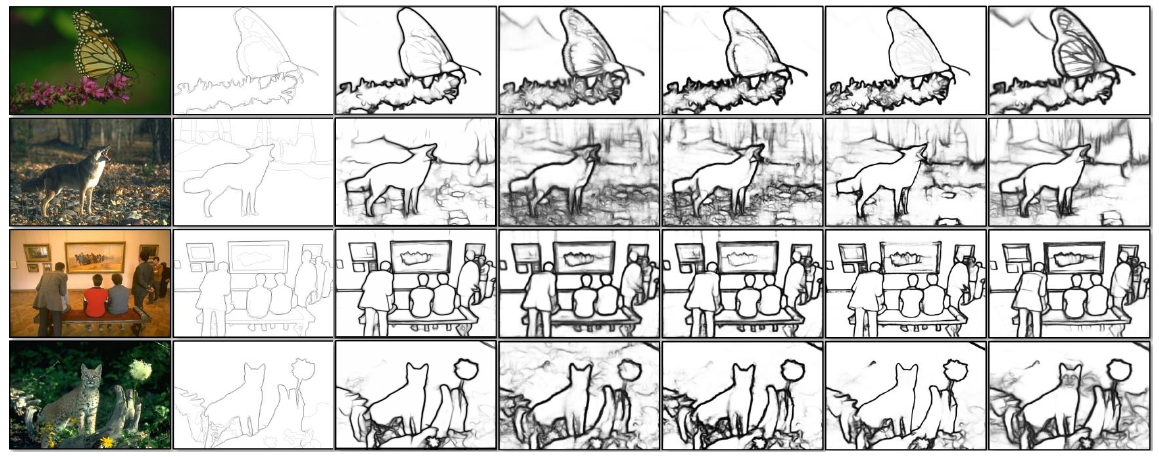}
    \hspace*{0.5cm}Image
    \hspace*{1.8cm}GT
    \hspace*{1.8cm}Ours
    \hspace*{1.6cm}RCF
    \hspace*{1.8cm}BDCN
    \hspace*{1.2cm}EDTER
    \hspace*{1.4cm}UAED
    \hspace*{0.5cm}
    
    \caption{Qualitative comparisons on BSDS dataset with previous state-of-the-arts.}
    \label{fig:bsds-vis}
\end{figure*}

We compare our model with traditional detectors (Canny~\cite{canny1986computational}, SCG~\cite{ren2012discriminatively}, OEF~\cite{hallman2015oriented}, \textit{etc}), CNN-based detectors (RCF~\cite{liu2019richer}, BDCN~\cite{he2022bdcn}, UAED~\cite{zhou2023uaed}, \textit{etc}) and the transformer-based detectors (EDTER~\cite{pu2022edter} and DiffusionEdge~\cite{ye2024diffusionedge}) on BSDS500. 
Quantitative results are shown in Table~\ref{tab:rsult-bsds}. By training on the trainval set of BSDS500, our NBED achieves the F-measure ODS of 0.834 with single-scale testing and obtains 0.840 with multi-scale testing, which is superior to all the state-of-the-art edge detectors. 

It's worth mentioning that NBED is a vanilla encoder-decoder with neither a complex training strategy nor a huge model size, which stands out among recent high performance edge detectors.
Among the detectors with similar performance to our NBED, EDTER~\cite{pu2022edter} not only has a model size of 11 times larger than NBED, but also requires multi-stage training. Nonetheless, the single-scale ODS of EDTER is still 1.4\% lower than that of NBED. The number of parameters of UAED~\cite{zhou2023uaed} is about twice that of NBED, and it requires multi-label data. The single-scale ODS of UAED in this case is 0.9\% lower than that of NBED. DiffusionEdge~\cite{ye2024diffusionedge} appears to perform competitively, with a single-scale ODS 0.4\% lower than NBED's, and DiffusionEdge not only relies on large-scale models, but also requires multi-step sampling during inference, which is very inefficient.
In addition to these methods with complex training strategies, the top performing vanilla encoder-decoder is FCL~\cite{xuan2022fcl} with a single-scale ODS of 0.807, which is more than 3\% lower than NBED and is no longer a contender.
Some qualitative results are shown in Fig.~\ref{fig:bsds-vis}. We observe that the proposed NBED shows a clear advantage in prediction quality, both crisp and accurate.

The GFLOPs of NBED are also very impressive, not only much lower than EDTER and DiffusionEdge, but also lower than models with fewer parameters such as RCF and BDCN. This mainly thanks to the design of the bilateral encoder, so that the high-resolution location extraction branch has few channels, and the self-attention module with high computational complexity deals with low-resolution features.

\begin{figure*}[ht]
    \centering
    \includegraphics[width=1\linewidth]{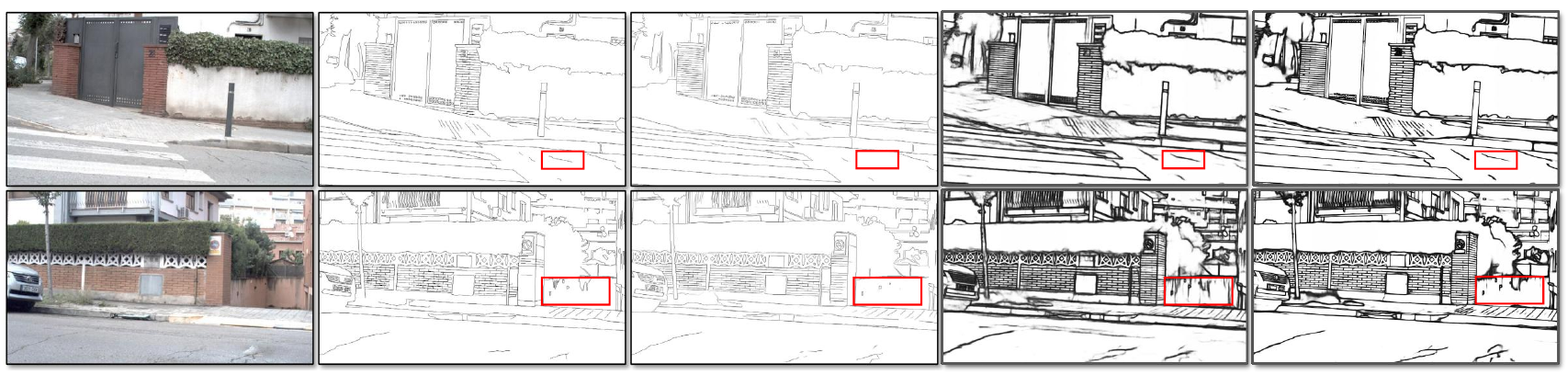}
    \hspace*{0.5cm}Image
    \hspace*{2.8cm}GT
    \hspace*{2.2cm}DiffusionEdge
    \hspace*{2cm}PiDiNet
    \hspace*{2.5cm}Ours\hspace*{0.5cm}
    
    \caption{Qualitative comparison on BIPED dataset with the previous state-of-the-art.}
    \label{fig:biped-vis}
\end{figure*}

To verify the reliability of our method NBED, we further experiment on NYUDv2 and BIPED datasets, and the experimental results are shown in table~\ref{tab:result-others}. 
We can clearly observe that NBED achieves competitive results on both datasets.
The ODS on NYUDv2 reaches 0.761, which is second only to EDTER and significantly better than other methods. 
The ODS on BIPED achieves 0.890, which is quite competitive with the most advanced methods DiffusionEdge and EDTER.
Some qualitative results are shown in Fig.~\ref{fig:biped-vis}. 
We compare our NBED with the latest method DiffusionEdge and CNN-based PiDiNet, 
The overall visual quality of the edges generated by our UAED is between PiDiNet and DiffusionEdge, while NBED handles the details better. As shown in the red box, the results of NBED are more similar to the ground truth at complex edges with low contrast, and there are fewer missed edges.

The size of the model does not change with the change of the dataset, and the relative relationship of the efficiency of each model is basically unchanged when the image resolution does not change much, so we do not repeat the model size and GFLOPs on NYUDv2 and BIPED, which can be referred to Table~\ref{tab:rsult-bsds}.

\begin{table}
\setlength{\tabcolsep}{4pt} 
    \centering
    \begin{tabular}{c|c|cc|cc}
    \toprule
    \multirow{2}*{Methods}& \multirow{2}*{Pub.'Year}& \multicolumn{2}{c|}{NYUDv2} & \multicolumn{2}{c}{BIPED} \\
         \cline{3-6}
         && ODS & OIS &ODS & OIS \\
         \hline
         AMH-Net& NeurIPS'17 & 0.744& 0.758 &-&-  \\
         HED& PAMI'17  &  0.722 & 0.737& 0.829 & 0.847   \\
         RCF& PAMI'19  &  0.745& 0.759&  0.843 &0.859    \\
         BDCN& PAMI'22  &  0.748 &0.762&  0.839 &0.854   \\
         DexiNed& PR’23 &-      &-     &0.859  &0.867    \\
         PiDiNet & ICCV'21  & 0.733 & 0.747 & 0.868& 0.876   \\
         EDTER&CVPR'22   &  \textbf{0.774} & \textbf{0.789}&  0.893& 0.898  \\
         DiffusionEdge& AAAI'24   &  0.761 & 0.766& \textbf{0.899}& \textbf{0.901}  \\
        NBED (ours)& - & 0.765&0.772 & 0.890 & 0.897 \\
    \bottomrule
    \end{tabular}
    \caption{Quantitative comparisons on NYUDv2 and BIPED. All results are computed with a single scale input.}
    \label{tab:result-others}
\end{table}

\subsection{Ablation study}

\begin{figure*}[ht]
    \centering
    \includegraphics[width=\linewidth]{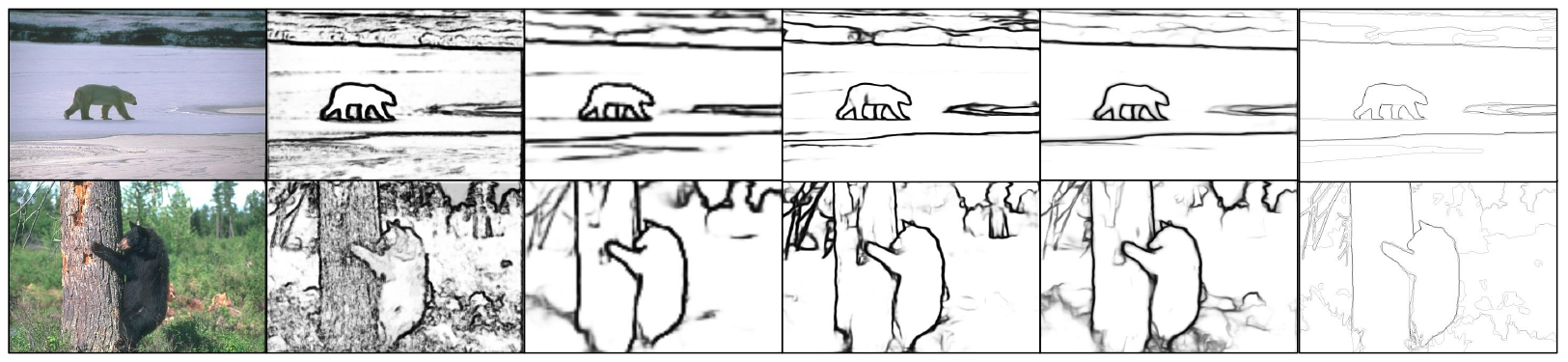}
    \hspace*{-0.5cm}Image
    \hspace*{1.5cm}Local Feature
    \hspace*{.6cm}Semantic Feature 
    \hspace*{1cm}VGG16 
    \hspace*{1.4cm}Our method
    \hspace*{1.5cm}GT
    \caption{Qualitative comparison of different encoders}
    \label{fig:abl-encoder-vis}
\end{figure*}

\begin{table}[ht]
    \centering
    \begin{tabular}{c|c|c|c|c}
    \toprule
         Encoder& Decoder & Modules&ODS & OIS\\
         \hline
         VGG16&  \multirow{4}*{Full}&\multirow{4}*{-}&0.797  &0.813 \\
         Local-Feat&&&0.688&0.703\\
         Seman-Feat && &0.823  &0.832 \\
         Full && &\textbf{0.838}  & \textbf{0.849}\\
         
         \hline
         \multirow{3}*{Full}&HED-style&\multirow{3}*{-}&0.830&0.841\\
         &UNet-style&&0.835&0.842\\
         &Full & &\textbf{0.838}  & \textbf{0.849}\\
         
         \hline
        \multirow{4}*{Full}&\multirow{4}*{Full}&CA&0.832&0.844\\
        &&ASPP&0.833&0.843\\
        &&CoFusion&\textbf{0.838}&0.847\\
        &&- &\textbf{0.838}  & \textbf{0.849}\\
         
    \bottomrule
    \end{tabular}
    \caption{Ablation study of the effectiveness of the components of the NBED on BSDS dataset. All results are computed with a single scale test. \textbf{\textit{Local-Feat}}: only use location feature, \textbf{\textit{Seman-Feat}}: only use semantic feature, \textbf{\textit{Full}}: Our proposed method (location feature + semantic feature), \textbf{\textit{CA}}: Channel Attention~\cite{su2021pixel}, \textit{\textbf{ASPP}}: Atrous Spatial Pyramid Pooling~\cite{he2022bdcn}, \textit{\textbf{CoFusion}}: Context-aware fusion block~\cite{huan2021unmixing}. }
    \label{tab:result-abl}
\end{table}

In ablation experiments, we verify the effectiveness of the components of our model, and the feature enhancement and fusion modules proposed in previous works, respectively. The experimental results are shown in Table~\ref{tab:result-abl}.

In the first set of experiments, we test the effect of various encoders. VGG16 is a widely used encoder in edge detection. when VGG16 is used as the encoder, the ODS of the model is 0.797, which is similar to RCF~\cite{liu2019richer} and LDC~\cite{soria2022ldc}. When only the location feature encoder of NBED is used, the ODS of the model is only 0.688, which is far worse than 0.823, the ODS semantic feature encoder. 
And these three encoders are far less effective than our NBED, which can achieve an ODS of 0.838.
A more intuitive difference can be found from Fig.~\ref{tab:result-abl}. 
The location feature encoder can preserve the location information of the edge, but due to the lack of semantic information guidance, the edge map is mixed with a lot of noise.
On the contrary, the semantic feature encoder can well suppress the expression of noise by using semantic information, but the lack of accurate location information produces aliasing in the upsampling process, which makes the edges rough. 
The results of VGG16 encoder look a little better, which is due to the ability of VGG16 to extract both location and semantic information. However, due to the lack of feature extraction ability of VGG16, it is difficult to meet the needs of edge detection, and ODS is much lower than NBED.
Another point worth noting is that although the edges generated by semantic feature encoder are not as good as VGG16 in visualizing, it is more accurate after NMS.

In the second set of experiments, we test the effect of various decoders. 
We compare our proposed decoder with the most commonly used HED-style decoder and UNet-style decoder. 
We can see that overall the decoder has less impact on the model performance than the encoder. In other words, what really matters in edge detection task is high-quality multi-scale features. If the quality of the multi-scale features is high, even a simple decoder like HED-style can achieve good performance, and the ODS of the model can achieve 0.830, which is better than most existing models. If the location error of semantic features in the HED-style decoder can be solved, the ODS of the model can reach 0.835 of UNet-style. On this basis, the expression of noise in shallow features is further suppressed, and the ODS of the model can be further improved to 0.838, which is the accuracy of our NBED.

In addition, we verify the effectiveness of the feature enhancement and fusion modules in previous methods~\cite{he2022bdcn,su2021pixel,huan2021unmixing} on our NBED. When the CA module and ASPP module are added, the accuracy of the model does not increase but decreases, which indicates that when the encoder is powerful enough, these additional feature enhancement modules not only cannot further enhance the features, but will have a negative impact. Using the feature fusion module CoFusion, the model accuracy is almost unchanged, which indicates that the shallowest features of NBED have obtained sufficient semantic information, and no additional gains can be obtained by fusing features at other scales.

\section{Conclusion and limitation}

In this paper, we propose NBED, a vanilla encoder-decoder based edge detector to avoid large model size and complex training strategies. We reduce the complexity of the encoder by decoupling the extraction process of location features and semantic features. And the semantic feature feature is used to refine the location feature step by step to suppress the noise in the location feature and the location error in the semantic feature. Our experimental results show that NBED is still quite competitive even compared to state-of-the-art methods. More importantly, the compact model and simple training strategy make NBED friendly for downstream tasks and industrial deployment.

\textbf{Limitations.} NBED does not make some targeted improvements for edge detection, therefore, NBED should theoretically be applicable to all pixel-level classification tasks, such as salient detection, semantic segmentation, crowd counting, \textit{etc}. Therefore, we will try more tasks in future work. 
Secondly, we rely on the powerful feature extraction and fusion capabilities of the model to alleviate the adverse impact caused by the problems existing in edge detection itself (such as label errors, unbalanced samples, etc.). 
It is worth discussing whether further addressing these issues can further improve the performance of NBED. 
Finally, the edges generated by NBED are far from the single-pixel width, and it is difficult to get rid of the dependence on post-processing. 
Finer edges is still a promising future direction to explore.

\section*{Declaration of competing interest}
The authors confirm that they have no financial or interpersonal conflicts that could have influenced the research presented in this study.











\printcredits

\bibliographystyle{cas-model2-names}

\bibliography{main}



\end{document}